\documentclass[journal]{IEEEtran} 
\usepackage{algorithm}
\usepackage{algpseudocode}
\usepackage{array}
\usepackage[caption=false,font=normalsize,labelfont=sf,textfont=sf]{subfig}
\usepackage{textcomp}
\usepackage{stfloats}
\usepackage{url}
\usepackage{verbatim}
\usepackage{graphicx}
\usepackage{cite}
\usepackage{amsmath}
\usepackage{amssymb}
\usepackage{xcolor}
\usepackage{booktabs}
\usepackage{multirow}
\usepackage{bbding}
\usepackage{orcidlink}
\usepackage{bbm}
\usepackage[table]{xcolor}
\usepackage[capitalise]{cleveref}

\begin{document}

\title{HSCP: A Two-Stage Spectral Clustering Framework for Resource-Constrained UAV Identification}
\author{Maoyu Wang\orcidlink{0009-0008-0633-6716}, Yao Lu\orcidlink{0000-0003-0655-7814}, ~\IEEEmembership{Student Member, IEEE}, Bo Zhou, Zhuangzhi Chen, Yun Lin\orcidlink{0000-0003-1379-9301},~\IEEEmembership{Senior Member, IEEE}, Qi Xuan\orcidlink{0000-0002-6320-7012},~\IEEEmembership{Senior Member,~IEEE}, Guan Gui,~\IEEEmembership{Fellow,~IEEE}
\thanks{This work was partially supported by the Key R\&D Program of Zhejiang under Grant 2022C01018 and by the National Natural Science Foundation of China under Grant U21B2001, 62301492 and 61973273. (Corresponding author: Qi Xuan)}
\thanks{Maoyu Wang is with the College of Computer science and Technology, Zhejiang University of Technology, Hangzhou, China (e-mail: wangmy.zjut@gmail.com)}
\thanks{Yao Lu, Zhuangzhi Chen, Qi Xuan are with the Institute of Cyberspace Security, College of Information Engineering, Zhejiang University of Technology, Hangzhou 310023, China, also with the Binjiang Institute of Artificial Intelligence, Zhejiang University of Technology, Hangzhou 310056, China (e-mail: yaolu.zjut@gmail.com, xuanqi@zjut.edu.cn, zzch@zjut.edu.cn).}
\thanks{Bo Zhou is with the Department of Intelligent Control, Zhejiang Institute of
Communications, Hangzhou 311112, China, and also with the ocean college,
Zhejiang University, Zhoushan 316021, and also with UniTTEC Co., Ltd,
Hangzhou 310051, China (e-mail: wxjs201@163.com).}
\thanks{Yun Lin is with the College of Information and Communication Engineering, Harbin Engineering University, Harbin, China (e-mail: linyun@hrbeu.edu.cn).}
\thanks{Guan Gui is with the College of Telecommunications and Information Engineering, Nanjing University of Posts and Telecommunications, Nanjing 210003, China (e-mail: guiguan@njupt.edu.cn).}}

\markboth{Journal of \LaTeX\ Class Files,~Vol.~14, No.~8, August~2021}%
{Shell \MakeLowercase{\textit{et al.}}: A Sample Article Using IEEEtran.cls for IEEE Journals}


\maketitle

\begin{abstract}
With the rapid development of Unmanned Aerial Vehicles (UAVs) and the increasing complexity of low-altitude security threats, traditional UAV identification methods struggle to extract reliable signal features and meet real-time requirements in complex environments. Recently, deep learning based Radio Frequency Fingerprint Identification (RFFI) approaches have greatly improved recognition accuracy. However, their large model sizes and high computational demands hinder deployment on resource-constrained edge devices. While model pruning offers a general solution for complexity reduction, existing weight, channel, and layer pruning techniques struggle to concurrently optimize compression rate, hardware acceleration, and recognition accuracy. To this end, in this paper, we introduce HSCP, a Hierarchical Spectral Clustering Pruning framework that combines layer pruning with channel pruning to achieve extreme compression, high performance, and efficient inference. In the first stage, HSCP employs spectral clustering guided by Centered Kernel Alignment (CKA) to identify and remove redundant layers. Subsequently, the same strategy is applied to the channel dimension to eliminate a finer redundancy. To ensure robustness, we further employ a noise-robust fine-tuning strategy. Experiments on the UAV-M100 benchmark demonstrate that HSCP outperforms existing channel and layer pruning methods. Specifically, HSCP achieves $86.39\%$ parameter reduction and $84.44\%$ FLOPs reduction on ResNet18 while improving accuracy by $1.49\%$ compared to the unpruned baseline, and maintains superior robustness even in low signal-to-noise ratio environments.
\end{abstract}


\section{Introduction}

\IEEEPARstart{W}{ith} the rapid advancement of precision sensor technologies and intelligent control systems, the application of Unmanned Aerial Vehicles has quickly gained global widespread~\cite{nemer2021rf}. Initially, UAVs were primarily deployed in the military domain, serving critical roles in reconnaissance, surveillance, and battlefield support~\cite{michalska2018capabilities,krishnan2015mass}. Today, their utility has expanded substantially into the civilian sector, encompassing aerial photography, logistics transportation, agricultural monitoring, and emergency rescue~\cite{yin2023joint,otto2018optimization,tsouros2019review}. These advancements have not only fueled rapid industrial growth but also brought convenience to social production and daily life. However, the widespread adoption of UAVs has concurrently given rise to potential security threats, including unauthorized intrusion into restricted airspace, illegal transport of goods, and risks to aviation safety~\cite{altawy2016security,levin2017faa}. Consequently, establishing a timely, accurate, and reliable UAV identification mechanism has become urgent for safeguarding public safety and enhancing low-altitude airspace management.

Current UAV detection methods are constrained by inherent limitations. For example, acoustic detection suffers from environmental noise interference~\cite{zhang2018deep}. Visual and optical methods, while intuitive, struggle with adaptive detection under complex background conditions, visual occlusion, and varying lighting conditions~\cite{gokcce2015vision,xie2020adaptive}. Radar-based detection can be compromised by obstructions such as buildings in urban environments, leading to blind spots and high deployment costs~\cite{guvencc2017detection,hommes2016detection}. In contrast, Radio Frequency Fingerprint Identification (RFFI) offers a distinct advantage by leveraging the unique hardware imperfections in wireless transmitters~\cite{danev2012physical}. These physical features, arising from minor variations in the manufacturing process of components like antennas and circuits, act as a unique fingerprint for each device. Since these characteristics are extremely difficult to forge or copy, RFFI has been widely utilized to distinguish between devices of the same model, providing a reliable means of identification in non-visual and adversarial environments~\cite{ezuma2019micro,al2019rf}.

In practical RFFI deployment, the constraints of real-world application sce narios must be rigorously addressed. Recognition systems are typically deployed on resource-limited edge devices, where computational capability and power consumption are critical bottlenecks~\cite{vasishta2024small}. Although existing deep learning models~\cite{jiang2025traditional,wang2025target,wang2024scantd,jin2025reasoning,ou2025social,ou2025application,ou2025analyzing,zhang2025time,sha2022acoustic,sha2022multi} have achieved impressive performance, they often contain a large number of parameters, resulting in significant inference delays that hinder real-time recognition. Furthermore, raw I/Q signals collected in practice are frequently corrupted by Gaussian noise, multipath effects, and time-varying channel conditions, necessitating that the model maintains robustness under noisy inputs~\cite{o2018over}. In time-critical low-altitude confrontation scenarios, both latency and reliability are decisive factors for control effectiveness. Consequently, achieving high recognition accuracy while significantly reducing model complexity and inference latency remains an urgent challenge.

Current mainstream studies primarily focus on enhancing recognition performance by designing new lightweight networks from scratch~\cite{zhou2025lightweight,cai2024toward,zhu2025lightweight,cai2024toward2}. However, such methods usually rely heavily on extensive architecture search and manual tuning, which are time-consuming and suffer from limited generalizability across diverse tasks. Other approaches attempt to improve model robustness against noise through complex data augmentation or training strategies, yet these do not fundamentally alleviate the burden of model parameters or inference costs. Overall, most existing methods are tailored for specific scenarios and lack a universal model compression framework applicable across mainstream architectures. This limitation results in high costs and significant barriers for engineering deployment and model migration.

To address these challenges and avoid the difficulty of manual design, we propose \textbf{HSCP}, which stands for \textbf{H}ierarchical \textbf{S}pectral \textbf{C}lustering \textbf{P}runing. This is a general and efficient method that significantly reduces parameters and latency without changing the overall architecture. Unlike methods that design new networks from scratch, our approach works directly on existing pre-trained models. It adapts to many common architectures such as ResNet~\cite{he2016deep}, ShuffleNet~\cite{ma2018shufflenet}, and MobileNet~\cite{sandler2018mobilenetv2}. Sepecifically, HSCP operates in a coarse-to-fine manner. In the first stage, we employ spectral clustering based on CKA similarity matrix to identify and remove redundant layers. Subsequently, we extend this unified analysis to the channel dimension to eliminate a finer redundancy. To ensure the model remains strong in noisy conditions, we add random Gaussian noise during the fine-tuning stage. This allows the pruned model to keep high accuracy across different noise levels. Compared with building new models, our method uses pre-trained networks to compress and accelerate the model at the same time. This reduces engineering work, speeds up edge deployment, and provides a useful solution for UAV situational awareness.

In summary, this work makes the following contributions:

\begin{itemize}
    \item \textbf{Hierarchical Spectral Clustering Pruning Implementation:} We propose HSCP, a hierarchical spectral clustering pruning framework that integrates layer and channel pruning. Specifically, we employ a unified spectral clustering strategy based on CKA matrix to identify and remove redundant layers and channels sequentially. This approach systematically reduces model complexity while preserving the capacity of the original network.

    \item \textbf{Noise-robust Fine-tuning Strategy:} When deployed in complex wireless environments, the performance of the pruned model will reduce due to signal interference and noise. To solve this problem we introduce a noise-robust fine-tuning strategy via Mixup augmentation. This technique generates virtual samples to enhance model generalization, ensuring the pruned model maintains high recognition accuracy even in low-SNR environments.

    \item \textbf{Universality and Effectiveness Validation:} Extensive experiments demonstrate that HSCP outperforms existing pruning methods across the UAV-M100 dataset and 3 mainstream architectures (ResNet, MobileNet, ShuffleNet) while achieving an optimal balance between parameter, compression and detection accuracy. Notably, our method shows superior robustness and universality compared to state-of-the-art baselines.

\end{itemize}





In the remainder of this paper, we first review the related work and introduce the prior knowledge in \cref{sec: Related Work} and \cref{sec:prior}. In \cref{sec: Method}, we detail the pruning framework. \cref{sec:Experiments} presents the experimental evaluation and analysis, followed by the conclusion in \cref{sec:Conclusion}.

\section{Related Work}
\label{sec: Related Work}
\subsection{UAV Identification} 

UAV identification has gained widespread attention as it supports airspace security and monitoring. Recent studies have explored multiple sensing modalities, including vision, radar, and radio frequency (RF) signals. Vision-based approaches leverage deep learning for UAV detection and tracking. However, their performance significantly degrades under occlusion or low illumination~\cite{zhu2018vision}. Radar-based methods utilize micro-Doppler signatures to distinguish UAV types, and several works have enhanced feature extraction using algorithms like Orthogonal Matching Pursuit~\cite{molchanov2014classification,luo2013micro}. Nevertheless, radar systems often require costly equipment and have limited deployment flexibility.

In contrast, RFFI has gained popularity due to its passive sensing and anti-spoofing capability. Early works extracted handcrafted physical-layer features, whereas recent methods employ deep learning to capture device-specific RF fingerprints~\cite{ezuma2019micro}. For example, a complex-valued convolutional neural network~\cite{zha2021specific} was proposed to identify UAVs from communication signals, while AirID·\cite{mohanti2020airid} injected custom RF fingerprints to enhance recognition robustness. Related studies on aircraft signal classification have also shown the potential of RF fingerprinting with few labeled samples~\cite{wang2022few}.

Although deep learning-based RFFI methods have achieved impressive results in signal classification taskss~\cite{shen2021radio,zhang2021radio,he2023radio}, the models these methods used exhibit high computational complexity and large model size. This leads to slow inference speeds and heavy resource consumption, hindering their practical deployment on resource-constrained edge devices. To address this challenge, researchers began to focus on designing lightweight models. For instance, some studies utilize compact architectures to reduce parameter counts while maintaining reasonable accuracy~\cite{zhou2025lightweight,cai2024toward,zhu2025lightweight,cai2024toward2}. However, these methods also have some limitations. Designing new networks from scratch is time-consuming and often lacks the generalization capability of large pre-trained models. This highlights the necessity of exploring a unified pruning framework that can systematically eliminate redundancy at multiple levels to achieve an optimal balance between efficiency and accuracy.

\subsection{Model Pruning} 
Model pruning has emerged as one of the most effective approaches for model compression. Existing pruning methods can generally be categorized into weight pruning~\cite{han2015learning,molchanov2016pruning,lee2018snip}, channel pruning~\cite{li2025sepprune,lin2020hrank}, and layer pruning~\cite{lu2024reassessing,tang2023sr,lu2024generic,lu2025structural,lu2022understanding,chen2018shallowing}.

Specifically, unstructured weight pruning focuses on zeroing out individual connections based on specific criteria, such as magnitude~\cite{han2015learning} or gradient information~\cite{molchanov2016pruning,lee2018snip}. Early works demonstrated that removing parameters with small absolute values could achieve high sparsity ratios without significant accuracy loss. Although these methods can theoretically reduce model size significantly, they produce irregular sparse matrices that require specialized hardware or libraries to realize actual speedups~\cite{han2016eie,wen2016learning}. On standard hardware platforms, the overhead of sparse matrix indexing often negates the computational gains.

In contrast, channel pruning improves inference speed by removing specific channels from each layer. For instance, Li et al.~\cite{li2016pruning} utilize L1-norm criteria to prune entire filters, while Network Slimming~\cite{liu2017learning} imposes sparsity-inducing regularization on batch normalization scaling factors. Subsequently, data-driven approaches like HRank~\cite{lin2020hrank} introduce feature map rank as a more robust importance metric. More recently, DepGraph~\cite{fang2023depgraph} explicitly models the dependency between coupled channels to handle complex architectures. While channel pruning is hardware-friendly and effectively reduces FLOPs, aggressive width reduction can severely damage the model's representational capacity, leading to a sharp decline in accuracy before achieving the desired compression rate.

Layer pruning selects those layers that contribute minimally to performance for removal. The concept originated from Stochastic Depth~\cite{huang2016deep}, which randomly dropped layers during training. Recent studies, such as Deep GreenAI~\cite{pons2025deep}, utilize block-wise similarity to identify and skip redundant layers. Layer pruning is particularly effective in reducing inference latency, as it directly shortens the forward propagation path. However, altering the network topology can disrupt signal propagation and gradient flow, making it challenging to maintain convergence and accuracy, especially for compact networks.

Although these pruning methods have achieved good results, each has its own limitations. Weight pruning is parameter-efficient but deployment-inefficient; channel pruning reduces memory but hits an accuracy bottleneck; and layer pruning significantly lowers latency but is sensitive to topological changes. For UAV detection tasks, the constraints are multifaceted: limited onboard memory requires model size reduction, while real-time processing demands low latency. To address this limitation, we propose a coarse-to-fine pruning framework that synergistically combines layer and channel pruning. By first removing redundant layers to reduce depth and subsequently pruning redundant channels to reduce width, our approach breaks through the compression ceiling of single-method strategies.

\section{prior knowledge}
\label{sec:prior}
\subsection{Dataset Details}
The dataset~\cite{soltani2020rf} contains RF signals from seven identical DJI M100 UAVs, captured in an RF anechoic chamber using an Ettus USRP X310 with a UBX $160$ daughterboard. Each UAV was flown individually at four distances ($6$, $9$, $12$, and $15$ ft) from the receiver while transmitting in a $10$ MHz downlink channel. At each distance, signals were collected in four 2-second bursts, separated by 10-second intervals. Each burst was divided into $140$ non-overlapping sequences of data, resulting in a dataset of more than 13,000 sequences, each with an average length of $92,000$ I/Q samples. Formally, the received composite signal from the k-th UAV is represented as
\begin{equation}
\begin{aligned}
y_k(t) = s_k(t) * c_k(t) + w_k(t), \quad k = 1,2,\dots,7
\end{aligned}
\end{equation}
where $y_k(t)$ corresponds to the signal received at the receiver from the $k$-th UAV, $s_k(t)$ indicates the UAV's transmitted signal,$*$ denotes the convolution operation, $c_k(t)$ accounts for the effects of the channel between the transmitter and receiver, and $w_k(t)$ represents the environment noise.

\begin{figure}[t]
	\centering
    \centering
    \includegraphics[width=\columnwidth]{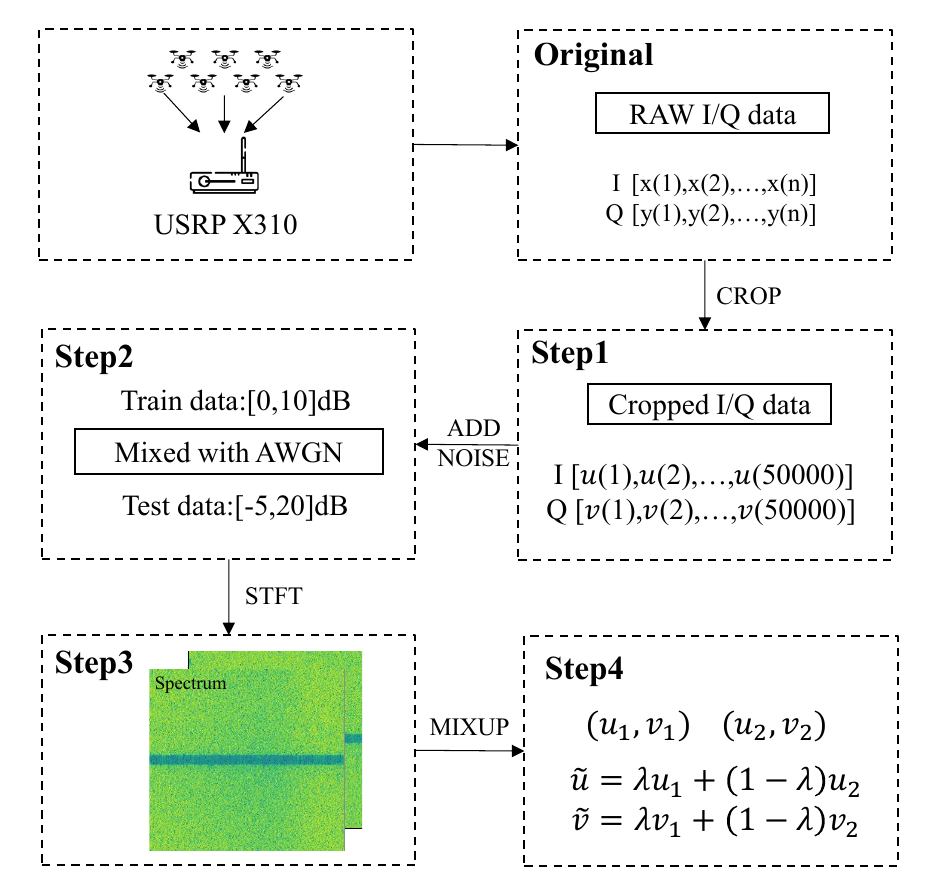} 
    \caption{Schematic of the data processing pipeline. Raw signals are processed via normalization, noise injection, and STFT transformation, followed by Mixup augmentation.} 
    \label{fig:dataset}
\end{figure}

\subsection{Data Processing}
The raw I/Q data require preprocessing before being input into the neural network. During data collection, UAV RF signals were captured using a USRP X310 and stored in IQ format. The overall preprocessing pipeline is illustrated in ~\cref{fig:dataset}.

\subsubsection{Data Normalization}
The captured I/Q samples have a wide range of sizes, from 79 KB to 840 KB, corresponding to approximately 215,000 points for the largest sample. Samples smaller than 200 KB are removed to retain sufficient signal information. All remaining sequences are truncated to a fixed length of 50,000 points to standardize the dataset.
\subsubsection{Noise Injection}
To emulate realistic wireless conditions, additive white Gaussian noise (AWGN) was introduced into the collected signals. Training data are processed with noise of SNR ratios between $0$ and $10$ decibels, while test data are exposed to a wider range from $-5$ to $20$ decibels to evaluate model robustness under diverse channel conditions.
\subsubsection{Time-Frequency Transformation}
The preprocessed I/Q sequences are then transformed into time-frequency spectrograms using the short-time Fourier transform (STFT). Each signal sequence is denoted as $s[m]$, and a finite-length window function $w[m]$ is applied to isolate a segment of the signal for analysis. The discrete STFT is calculated as
\begin{equation}
X(t, f) = \sum_{m=0}^{L-1} s[m] \cdot w[m-t] \cdot e^{-j 2 \pi f m / L},
\end{equation}
where $t$ represents the time index, $f$ represents the frequency index, $L$ is the number of points in the transform, and the exponential term performs the Fourier transform at each frequency. The window function ensures that only a local segment of the signal contributes to the calculation at each time step, allowing the STFT to capture the time-varying frequency characteristics of the signal. The resulting complex matrix $X(t,f)$ contains both amplitude and phase information. The spectrogram is then expressed as
\begin{equation}
S(t, f) = \log_{10} |X(t, f)|^2.
\end{equation}
This representation emphasizes the signal energy distribution across time and frequency and reduces the influence of multipath fading, Doppler shifts, and other channel effects.

\subsubsection{Mixup Data Augmentation}
To further enhance model generalization and mitigate overfitting, Mixup data augmentation was applied. New samples were generated by linearly combining pairs of original signals and their labels. For two training samples $(u_1, v_1)$ and $(u_2, v_2)$, a new sample $(\tilde{u}, \tilde{v})$ is obtained as
\begin{equation}
\tilde{u} = \lambda u_1 + (1-\lambda) u_2, \quad
\tilde{v} = \lambda v_1 + (1-\lambda) v_2,
\end{equation}
where $\lambda$ is drawn from a Beta distribution. This approach increases the variability of the training data, allowing the model to learn smoother decision boundaries and adapt to unseen signals.

\begin{figure*}[t]
	\centering
    \centering
    \includegraphics[width=0.99\textwidth]{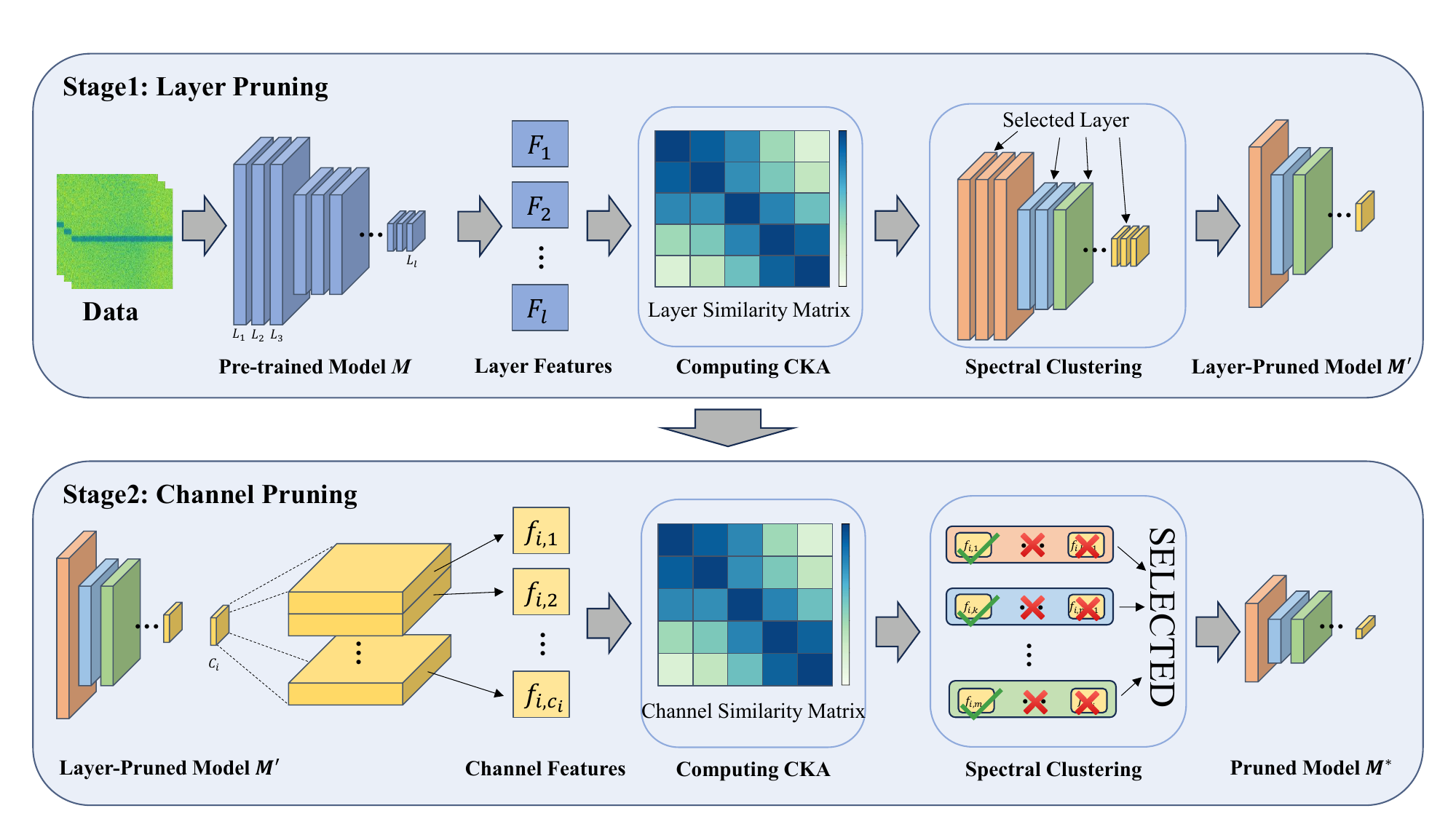} 
 \caption{Illustration of the two-stage pruning pipeline. The method sequentially performs layer and channel pruning using a unified CKA-based spectral clustering strategy. In both stages, only the leading components from each cluster are retained while redundant ones are removed, resulting the final structurally compressed model.} 
 \label{fig:method}
\end{figure*}

\section{Method}
\label{sec: Method}

In this section, we present the Hierarchical Spectral Clustering Pruning method \textbf{HSCP}, which performs layer pruning followed by channel pruning to achieve a compact yet high-performing model. Both layer pruning and channel pruning share a unified pruning strategy that measures the representational similarity between structural components and selectively removes redundant ones. Specifically, we first extract feature representations from the pre-trained model, compute the pairwise similarity using Centered Kernel Alignment (CKA), and then employ spectral clustering to group highly similar layers or channels. Within each cluster, we preserve the first few subset, while the remaining components are pruned. The schematic overview of this framework is illustrated in \cref{fig:method}.

\subsection{Layer Pruning}
In the first stage, we prune redundant layers to obtain a structurally compact model while maintaining task performance. Let the pre-trained network be denoted as $\mathbf{M} = L_1 \circ L_2 \circ \dots \circ L_l$, where $L_i$ represents the $i$-th layer and $\circ$ indicates function composition.

To measure redundancy between layers, we first extract the intermediate feature representations $\mathbf{F}_i$ for each layer using a batch of input samples $\mathbf{X}_b$:
\begin{equation}
    \mathbf{F}_i = L_1 \circ L_2 \circ \dots \circ L_i(\mathbf{X}_b), \qquad i = 1, \dots, l,
\end{equation}
where $\mathbf{F}_i \in \mathbb{R}^{b \times c_i \times h_i \times w_i}$ denotes the activation tensor with batch size $b$, channel number $c_i$, height $h_i$, and width $w_i$. For computational convenience, each activation map is flattened as $\hat{\mathbf{F}}_i \in \mathbb{R}^{b \times (c_i \times h_i \times w_i)}$. 

The statistical dependence between layers $i$ and $j$ is quantified using the Hilbert--Schmidt Independence Criterion (HSIC):
\begin{equation}
\begin{split}
    \text{HSIC}_1(\mathbf{K}, \mathbf{L}) = & \\
    \frac{1}{b(b-3)}
    \Big[
        \mathrm{tr}(\tilde{\mathbf{K}}\tilde{\mathbf{L}}) 
        &+ \frac{\mathbf{1}^\top \tilde{\mathbf{K}}\mathbf{1}\,\mathbf{1}^\top \tilde{\mathbf{L}}\mathbf{1}}{(b-1)(b-2)} 
        - \frac{2}{b-2}\mathbf{1}^\top \tilde{\mathbf{K}}\tilde{\mathbf{L}}\mathbf{1}
    \Big], 
\end{split}
\label{eq:hsic}
\end{equation}
where $\mathbf{K} = \hat{\mathbf{F}}_i \hat{\mathbf{F}}_i^\top$ and $\mathbf{L} = \hat{\mathbf{F}}_j \hat{\mathbf{F}}_j^\top$ are the Gram matrices of layer $i$ and $j$, respectively; $\tilde{\mathbf{K}} = \mathbf{K}(\mathbf{1}\mathbf{1}^\top - \mathbf{I}_b)$ and $\tilde{\mathbf{L}} = \mathbf{L}(\mathbf{1}\mathbf{1}^\top - \mathbf{I}_b)$ are their centered forms; $\mathbf{I}_b$ is the $b$-dimensional identity matrix and $\mathbf{1}$ denotes an all-ones vector. 

Because HSIC is sensitive to isotropic scaling, we normalize it using the Centered Kernel Alignment (CKA) metric:
\begin{equation}
    \text{CKA}(\mathbf{K}, \mathbf{L}) 
    = \frac{\text{HSIC}_1(\mathbf{K}, \mathbf{L})}
    {\sqrt{\text{HSIC}_1(\mathbf{K}, \mathbf{K}) \, \text{HSIC}_1(\mathbf{L}, \mathbf{L})}},\label{eq:cka}
\end{equation}
which yields a similarity score within $[0,1]$ between the two layers. By evaluating all layer pairs, we construct the symmetric similarity matrix:
\begin{equation}
    \mathbf{S} = [S_{ij}]_{l \times l}, \quad S_{ij} = \text{CKA}(L_i, L_j),
\end{equation}
where $S_{ij}$ indicates the representational similarity between layers $L_i$ and $L_j$. 

To identify groups of layers with similar functionality, we perform spectral clustering based on $\mathbf{S}$. Treating $\mathbf{S}$ directly as the affinity matrix, the algorithm constructs the normalized graph Laplacian:
\begin{equation}
    \mathbf{L}_{\mathrm{sym}} = \mathbf{I}_l - \mathbf{D}^{-1/2}\mathbf{S}\mathbf{D}^{-1/2}, \label{eq:laplacian_layer}
\end{equation}
where $\mathbf{I}_l$ is the $l \times l$ identity matrix and $\mathbf{D} \in \mathbb{R}^{l \times l}$ is the degree matrix with diagonal entries $\mathbf{D}_{ii} = \sum_{j=1}^{l} S_{ij}$. An eigen-decomposition of $\mathbf{L}_{\mathrm{sym}}$ is then performed:
\begin{equation}
    \mathbf{L}_{\mathrm{sym}} \mathbf{u}_i = \lambda_i \mathbf{u}_i, \qquad i = 1, \dots, l,\label{eq:lu}
\end{equation}
where $\lambda_i$ and $\mathbf{u}_i$ denote the $i$-th eigenvalue and corresponding eigenvector, respectively. The $k$ eigenvectors associated with the smallest nonzero eigenvalues are selected to form the embedding matrix
\begin{equation}
    \mathbf{U} = [\mathbf{u}_1, \mathbf{u}_2, \dots, \mathbf{u}_k] \in \mathbb{R}^{l \times k}.
\end{equation}
Each row $\mathbf{U}_{i,:}$ is normalized to unit length:
\begin{equation}
    \mathbf{Y}_{i,:} = \frac{\mathbf{U}_{i,:}}{\|\mathbf{U}_{i,:}\|_2}, \quad i = 1, \dots, l,
\end{equation}
and a $k$-means algorithm is applied to $\{\mathbf{Y}_{i,:}\}_{i=1}^l$ to obtain $k$ clusters of layers.

Each cluster $B_m$ ($m = 1, \dots, k$) corresponds to a block of layers exhibiting similar representation. Within each block, we retain the first layer as the representative while discarding the others, removing redundant transformations without relying on complex importance scores. The preserved layers are then reassembled into a compact model $\mathbf{M}'$. If dimensional mismatches arise between consecutive retained layers, the subsequent layer is reinitialized using Kaiming initialization to ensure shape consistency. The resulting pruned model $\mathbf{M}'$ is directly forwarded to the channel pruning stage for finer reduction.

\subsection{Channel Pruning}
After completing layer pruning, we further explore redundancy by pruning redundant channels in each remaining layer. This stage employs the same redundancy analysis framework as layer pruning but is applied to the channel dimension ($p, q$) instead of entire layers ($i, j$).

Given the activation tensor $\mathbf{F}_i \in \mathbb{R}^{b \times c_i \times h_i \times w_i}$ of the $i$-th retained layer, we first reshape it into $\hat{\mathbf{F}}_i = [\mathbf{f}_{i,1}, \dots, \mathbf{f}_{i,c_i}]^\top$, where each $\mathbf{f}_{i,p}$ denotes the flattened response of the $p$-th channel.  We then calculate the similarity score $S^{(i)}_{pq}$ between the $p$-th and $q$-th channels using the CKA metric as defined in \cref{eq:cka}, utilizing the HSIC formulation from \cref{eq:hsic}:
\begin{equation}
    S^{(i)}_{pq} = \text{CKA}(\mathbf{f}_{i,p}, \mathbf{f}_{i,q}).
\end{equation}
By calculating this for all channel pairs, we obtain a channel similarity matrix $\mathbf{S}^{(i)} \in \mathbb{R}^{c_i \times c_i}$ which captures the pairwise redundancy within the layer.

We then perform spectral clustering on $\mathbf{S}^{(i)}$ to partition the channels into $k_i$ clusters. The clustering process follows the same spectral relaxation logic as described in the layer pruning stage. Specifically, the normalized graph Laplacian $\mathbf{L}_{\mathrm{sym}}^{(i)}$ for channels is constructed using \cref{eq:laplacian_layer}:
\begin{equation}
    \mathbf{L}_{\mathrm{sym}}^{(i)} = \mathbf{I}_{c_i} - \mathbf{D}^{-1/2} \mathbf{S}^{(i)} \mathbf{D}^{-1/2},
\end{equation}
where $\mathbf{I}_{c_i}$ is the identity matrix and $\mathbf{D}$ is the degree matrix of $\mathbf{S}^{(i)}$. Subsequently, we perform the eigen-decomposition of $\mathbf{L}_{\mathrm{sym}}^{(i)}$ as in \cref{eq:lu} to obtain the spectral embeddings. The $k_i$ eigenvectors associated with the smallest eigenvalues are selected and clustered using $k$-means.

Channels grouped within the same cluster are identified as functionally redundant. To eliminate this redundancy, we retain the leading channel of each cluster as the representative and prune the remaining ones. Concurrently, the corresponding input filters in the subsequent convolutional layer are removed to align with the reduced channel dimension. This procedure yields the final structurally pruned model, denoted as $\mathbf{M}^*$. Finally, $\mathbf{M}^*$ undergoes a fine-tuning phase to recover its accuracy. We provide a summary of the HSCP pipeline in \cref{algo:pruning}.

\begin{algorithm}[t]
\caption{Hierarchical Spectral Pruning Framework}
\label{algo:pruning}
\textbf{Input:} A batch of examples $\mathbf{X}_b$, a pre-trained model $\mathbf{M} = L_1 \circ L_2 \cdots \circ L_l$, number of layer clusters $k$, and channel clusters $\{k_i\}$.\\
\textbf{Output:} The final pruned model $\mathbf{M}^*$.

\begin{algorithmic}[1]
\State \textbf{Stage 1: Layer Pruning}
\For{$i = 1$ to $l$}
    \State $\mathbf{F}_i = L_1 \circ L_2 \cdots \circ L_i(\mathbf{X}_b)$
    \State $\hat{\mathbf{F}}_i = \text{flatten}(\mathbf{F}_i)$
    \State $\mathbf{K} = \hat{\mathbf{F}}_i \hat{\mathbf{F}}_i^\top$
    \For{$j = 1$ to $l$}
        \State $\mathbf{F}_j = L_1 \circ L_2 \cdots \circ L_j(\mathbf{X}_b)$
        \State $\hat{\mathbf{F}}_j = \text{flatten}(\mathbf{F}_j)$
        \State $\mathbf{L} = \hat{\mathbf{F}}_j \hat{\mathbf{F}}_j^\top$
        \State Using Eq. (\ref{eq:cka}) with $\mathbf{K}$ and $\mathbf{L}$ to calculate $S_{ij}$.
    \EndFor
\EndFor
\State Using Eq. (\ref{eq:laplacian_layer}) to construct $\mathbf{L}_{\mathrm{sym}}$ from $\mathbf{S}$.
\State Using Eq. (\ref{eq:lu}) to obtain eigenvectors $\mathbf{U}$ and partition layers into $k$ blocks $\{B_m\}$.
\State Constructing $\mathbf{M}'$ by retaining the first layer of each block $B_m$.

\State \textbf{Stage 2: Channel Pruning}
\For{each layer $L_i$ in $\mathbf{M}'$}
    \State $\hat{\mathbf{F}}_i = [\mathbf{f}_{i,1}, \dots, \mathbf{f}_{i,c_i}]^\top$
    \For{$p = 1$ to $c_i$}
        \State $\mathbf{K} = \mathbf{f}_{i,p} \mathbf{f}_{i,p}^\top$
        \For{$q = 1$ to $c_i$}
            \State $\mathbf{L} = \mathbf{f}_{i,q} \mathbf{f}_{i,q}^\top$
            \State Using Eq. (\ref{eq:cka}) with $\mathbf{K}$ and $\mathbf{L}$ to calculate $S^{(i)}_{pq}$.
        \EndFor
    \EndFor
    \State Using Eq. (\ref{eq:laplacian_layer}) to construct $\mathbf{L}_{\mathrm{sym}}^{(i)}$ from $\mathbf{S}^{(i)}$.
    \State Using Eq. (\ref{eq:lu}) to partition channels into $k_i$ groups.
    \State Pruning channels by preserving the leading channel of each group.
    \State Removing corresponding input filters in the subsequent layer.
\EndFor
\State \textbf{Obtaining the final structurally pruned model $\mathbf{M}^*$.}
\end{algorithmic}
\end{algorithm}

\begin{table}[t]
  \centering
  \caption{Baseline Performance of the Original Unpruned Models on the UAV-M100 Dataset.}
    \begin{tabular}{cccc}
    \toprule
    Model & Acc (\%) & Params (M) & FLOPs (M) \\
    \midrule
    ResNet18 & 92.62 & 11.17 & 1552.33 \\
    MobileNet-V2 & 92.04 & 2.23  & 277.01 \\
    ShuffleNet-V2-1.0 & 88.58 & 1.26  & 134.72 \\
    \bottomrule
    \end{tabular}%
  \label{tab:original}%
\end{table}%

\section{Experiments}
\label{sec:Experiments}

\begin{table*}[htbp]
  \centering
    \caption{Pruning Results of Various Pruning Methods Applied to ResNet18, MobileNet-V2, and ShuffleNet-V2 on the UAV-M100 Dataset.}
    \resizebox{\textwidth}{!}{
    \begin{tabular}{c|c|ccccccccc}
    \toprule
    \multicolumn{1}{c}{Dataset} & \multicolumn{1}{c}{Model} & Method & Acc (\%) & $\Delta$Acc (\%) & Params (M) & $\Delta$Params (\%) & FLOPs (M) & $\Delta$FLOPs (\%) & Latency (ms) & GPU Memory \\
    \midrule
    \multirow{18}[6]{*}{UAV-M100} & \multirow{6}[2]{*}{ResNet18} & Random & 91.61 & -1.01 & 4.90   & 56.13 & 722.28 & 53.47 & 824.98 & 1852MB \\
          &       & HRank & 92.10  & -0.52 & 2.80   & 74.93 & 397.94 & 74.36 & 459.49 & 1668MB \\
          &       & Sr-init & 90.67 & -1.95 & 2.15  & 80.75  & 542.42 & 65.06 & 412.22 & 2168MB \\
          &       & PSR   & 91.61 & -1.01 & 2.10  & 81.20  & 533.96 & 65.60 & 334.66 & 2054MB \\
          &       & Ours  & 91.62 & -1.00    & 1.52  & 86.39 & 241.58 & 84.44 & 254.65 & 1386MB \\
          &       & \cellcolor[rgb]{ .867,  .922,  .969}Ours+Mixup & \cellcolor[rgb]{ .867,  .922,  .969}94.11 & \cellcolor[rgb]{ .867,  .922,  .969}1.49  & \cellcolor[rgb]{ .867,  .922,  .969}1.52  & \cellcolor[rgb]{ .867,  .922,  .969}86.39 & \cellcolor[rgb]{ .867,  .922,  .969}241.58 & \cellcolor[rgb]{ .867,  .922,  .969}84.44 & \cellcolor[rgb]{ .867,  .922,  .969}254.65 & \cellcolor[rgb]{ .867,  .922,  .969}1386MB \\
\cmidrule{2-11}          & \multirow{6}[2]{*}{MobileNet-V2} & Random & 91.27 & -0.77 & 1.01  & 54.71 & 84.98 & 69.32 & 513.30 & 3548MB \\
          &       & HRank & 90.67 & -1.37 & 1.16  & 47.98 & 182.61 & 34.08 & 1471.24 & 7224MB \\
          &       & Sr-init & 91.14 & -0.01  & 0.80  & 64.25 & 92.10 & -66.46 & 992.998 & 3860MB \\
          &       & PSR   & 92.51 & 0.47  & 0.85  & 61.88 & 118.71 & 57.15 & 780.78 & 4708MB \\
          &       & Ours  & 91.21 & -0.83 & 0.50   & 77.58 & 62.81 & 77.33 & 513.90 & 3534MB \\
          &       & \cellcolor[rgb]{ .867,  .922,  .969}Ours+Mixup & \cellcolor[rgb]{ .867,  .922,  .969}95.64 & \cellcolor[rgb]{ .867,  .922,  .969}3.60   & \cellcolor[rgb]{ .867,  .922,  .969}0.50   & \cellcolor[rgb]{ .867,  .922,  .969}77.58 & \cellcolor[rgb]{ .867,  .922,  .969}62.81 & \cellcolor[rgb]{ .867,  .922,  .969}77.33 & \cellcolor[rgb]{ .867,  .922,  .969}513.90 & \cellcolor[rgb]{ .867,  .922,  .969}3534MB \\
\cmidrule{2-11}          & \multirow{6}[2]{*}{ShuffleNet-V2-1.0} & Random & 83.92 & -4.66 & 0.75  & 40.48 & 53.32 & 60.42 & 258.42 & 1406MB \\
          &       & HRank & 88.20  & -0.38 & 0.80   & 36.51 & 82.44 & 38.81 & 591.88 & 2252MB \\
          &       & Sr-init & 90.89 & 2.31  & 0.76  & 39.68 & 76.30  & 43.36 & 259.30 & 1740MB \\
          &       & PSR   & 88.88 & 0.30   & 0.84  & 33.33 & 77.02 & 42.83 & 251.39 & 1726MB \\
          &       & Ours  & 86.97 & -1.61 & 0.26  & 79.37 & 27.99 & 79.22 & 265.66 & 1342MB \\
          &       & \cellcolor[rgb]{ .867,  .922,  .969}Ours+Mixup & \cellcolor[rgb]{ .867,  .922,  .969}92.25 & \cellcolor[rgb]{ .867,  .922,  .969}3.67  & \cellcolor[rgb]{ .867,  .922,  .969}0.26  & \cellcolor[rgb]{ .867,  .922,  .969}79.37 & \cellcolor[rgb]{ .867,  .922,  .969}27.99 & \cellcolor[rgb]{ .867,  .922,  .969}79.22 & \cellcolor[rgb]{ .867,  .922,  .969}265.66 & \cellcolor[rgb]{ .867,  .922,  .969}1342MB \\
    \bottomrule
    \end{tabular}%
    }
  \label{tab:overall}%
\end{table*}%

\subsection{Experiment Settings}
\label{sec:Experiment Settings}
We implement the deep learning components of our proposed framework using the PyTorch library and executed all computations on a NVIDIA A100 GPU platform. To perform the core structural analysis, we leverage the spectral clustering implementation from the Scikit-learn library which efficiently processes the affinity matrices based on layer-wise and channel-wise feature similarities. The experiments are conducted on the UAV-M100 dataset where input images are resized to a resolution of $102 \times 389$ pixels to accommodate standard network dimensions. For the optimization of both baseline and pruned models, we employ the Adam optimizer. The training process span $50$ epochs with a batch size of $64$ samples and we initialize the learning rate at $0.001$. During the fine-tuning phase of the pruned models, we incorporate the Mixup data augmentation strategy with the parameter $\alpha$ set to $0.5$ to enhance the generalization capability of the compact networks.

\subsection{Results and Analysis}
\label{sec:Results and Analysis}
We evaluate HSCP on the UAV-M100 dataset using three representative architectures: ResNet18~\cite{he2016deep}, MobileNet-V2~\cite{sandler2018mobilenetv2}, and ShuffleNet-V2~\cite{ma2018shufflenet}. \cref{tab:original} details the baseline performance of these unpruned models. For a comprehensive comparative analysis, we compare our method against several state-of-the-art (SOTA) pruning methods including Random pruning, HRank~\cite{lin2020hrank}, Sr-init~\cite{tang2023sr}, and PSR~\cite{lu2024generic}. The detailed quantitative comparisons of the pruned models are presented in \cref{tab:overall}.

\begin{figure}[t]
	\centering
    \centering
    \includegraphics[width=\columnwidth]{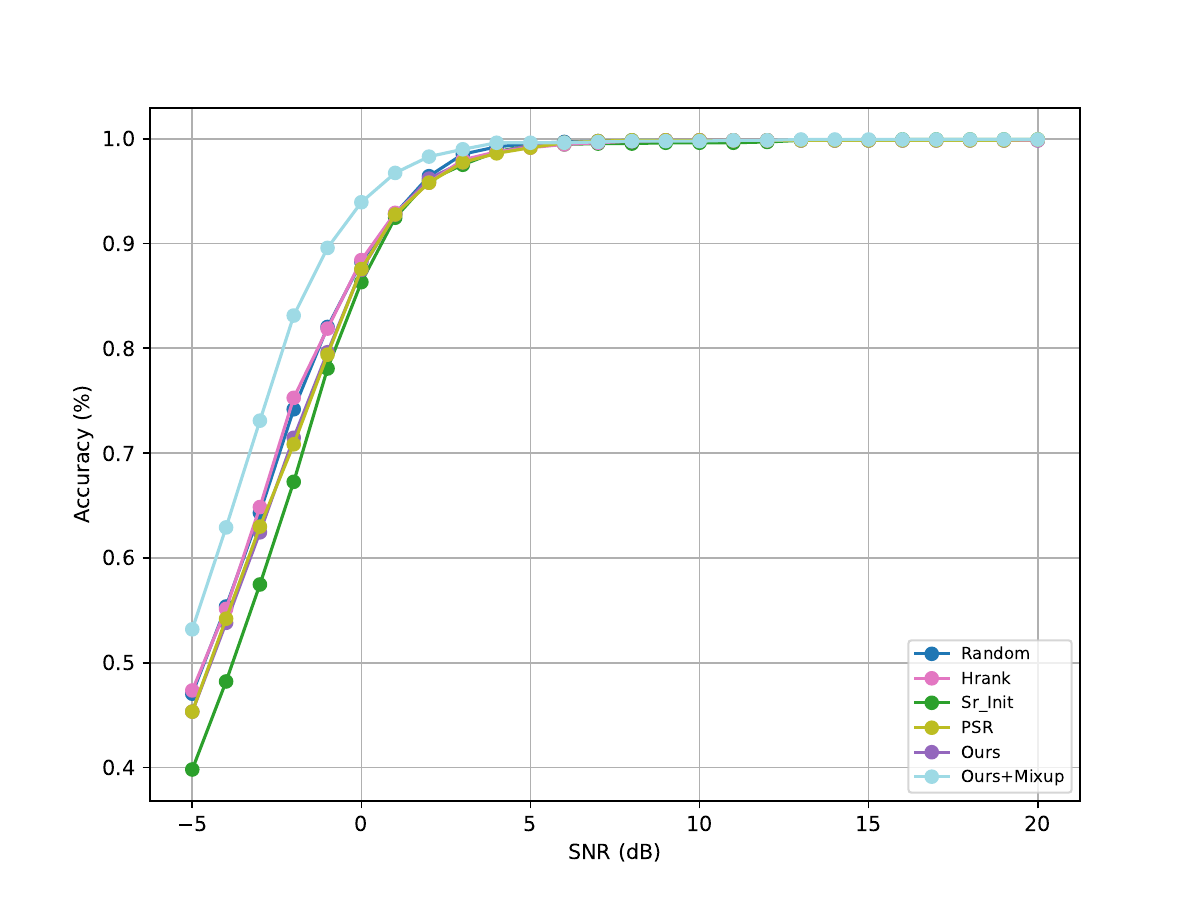} 
 \caption{Performance evaluation of pruned ResNet18 under varying SNR. The figure displays accuracy variation with SNR ranging from $-5$ dB to $20$ dB for the baseline methods and ours.}
 \label{fig:all_acc}
\end{figure}

\textbf{Performance on Standard Convolutional Networks.} The experimental results on ResNet18 demonstrate that HSCP achieves an exceptional trade-off between model compression and detection accuracy. As detailed in \cref{tab:overall}, our method aggressively reduces the model complexity by lowering the parameter count to 1.52 M and FLOPs to 241.58 M. This corresponds to reduction rates of $86.39\%$ and $84.44\%$, respectively. In comparison, the most competitive baseline, PSR, retains $2.10$M parameters and $533.96$M FLOPs. Consequently, our pruned model requires approximately $27.6\%$ fewer parameters and $54.7\%$ fewer FLOPs than the PSR method, highlighting the efficacy of our selection in identifying redundant layers and channels. Moreover, while aggressive pruning typically degrades performance, our method combined with the Mixup strategy achieves a top-1 accuracy of $94.11\%$. This represents a significant improvement of $1.49\%$ over the unpruned baseline and outperforms the PSR method by a margin of $2.50\%$. \cref{fig:all_acc} details the performance of ResNet18 with different pruning methods across a spectrum of SNR values. The visualized results reveal that while most methods converge to high accuracy in clean environments where SNR exceed $5$dB. Meanwhile our proposed framework demonstrates superior resilience under severe signal interference. Specifically, in the challenging low-SNR ranging from $-5$dB to $0$dB, the \textit{Ours+Mixup} consistently maintains the highest detection rates. For instance, at the extreme noise condition of $-5$dB, our method sustains an accuracy of approximately $53\%$ which significantly outperforms the Sr-init where performance drops to nearly $40\%$. This demonstrates that our method effectively extracts features and maintains superior robustness against signal perturbations.

\textbf{Robustness on Lightweight Architectures.} Pruning compact architectures is challenging due to limited redundancy. However, our method demonstrates the superiority over existing methods in terms of robustness. On the MobileNet-V2 architecture, previous methods such as HRank and PSR achieve FLOPs reductions of $34.08\%$ and $57.15\%$, respectively. In contrast, our approach significantly pushes the compression boundary to achieve a $77.33\%$ reduction in FLOPs and a $77.58\%$ reduction in parameters. Despite this high compression rate, \textit{Ours+Mixup} attains an accuracy of $95.64\%$, surpassing the closest competitor, Sr-init, by over $3\%$. Similarly, our method demonstrates the superiority on ShuffleNet-V2. While existing methods including HRank, Sr-init, and PSR delete approximately $40\%$ FLOPs reduction, our method breaks this bottleneck by achieving a $79.22\%$ reduction in FLOPs and reducing parameters to $0.26$M. This result is particularly notable as it offers three times greater parameter reduction than the PSR method which retains $0.84$M parameters. Furthermore, our model maintains a dominant accuracy of $92.25\%$, marking a $3.67\%$ increase over the baseline. This confirms that HSCP can effectively extract the most discriminative features even from highly lightweight networks.

\textbf{Inference Latency and Memory Efficiency.} The practical deployment value of HSCP is evidenced by its performance in inference latency and GPU memory usage. Unlike usual pruning methods that may introduce fragmented memory access, HSCP directly reduces network depth and leads to linear gains in speed. On ResNet18, our method reduces inference latency to $254.65$ms, achieving a speedup of approximately $3.2$ times compared to the baseline and significantly outperforming the $412.22$ms latency of Sr-init. A similar trend is observed on MobileNet-V2 where our method achieves a latency of $513.90$ms.  Furthermore, our method consistently consumes the least GPU memory across all models, exemplified by the $1386$MB usage for ResNet18 compared to $2054$MB for PSR. These results demonstrate that HSCP is the practical suitability for real-time and resource-limited UAV applications.

\begin{figure}[t]
	\centering
    \centering
    \includegraphics[width=\columnwidth]{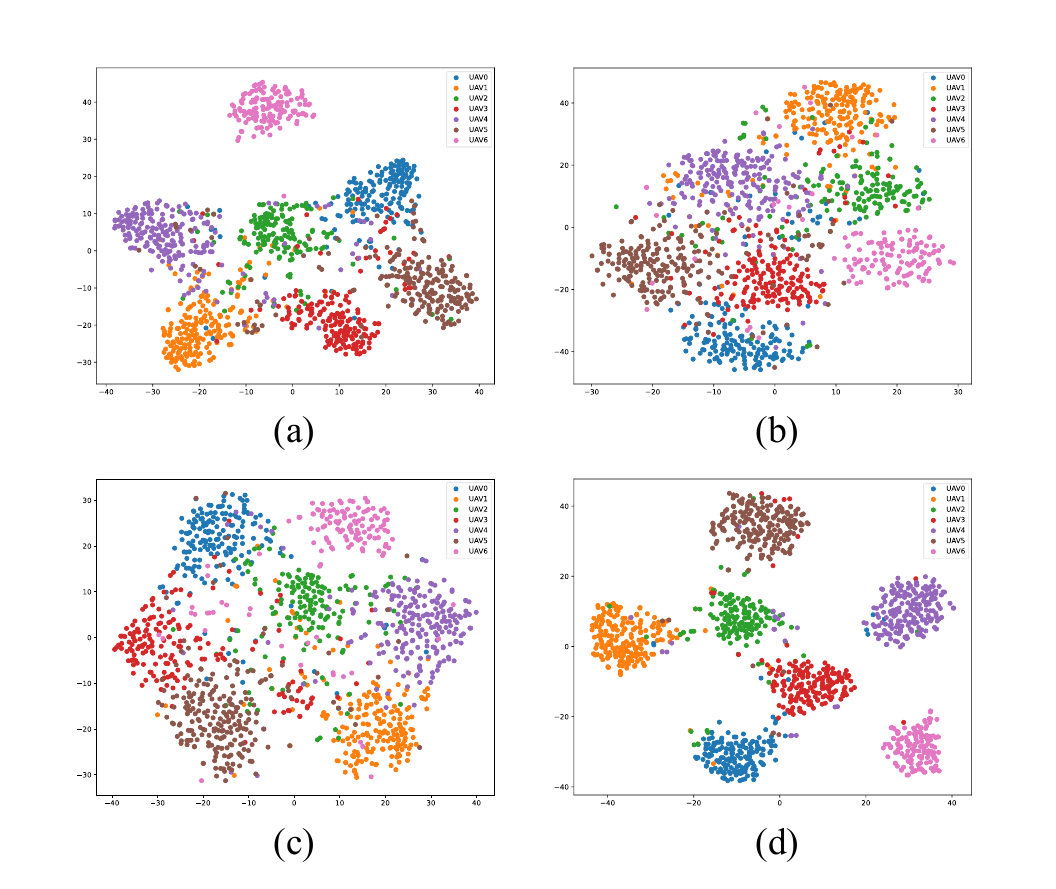} 
\caption{t-SNE visualization of feature representations extracted by different pruning models at an SNR of 0 dB. The subplots correspond to (a) HRank, (b) Sr-init, (c) PSR, and (d) Ours.}
\label{fig:tsne}
\end{figure}

\begin{figure}[t]
	\centering
    \centering
    \includegraphics[width=\columnwidth]{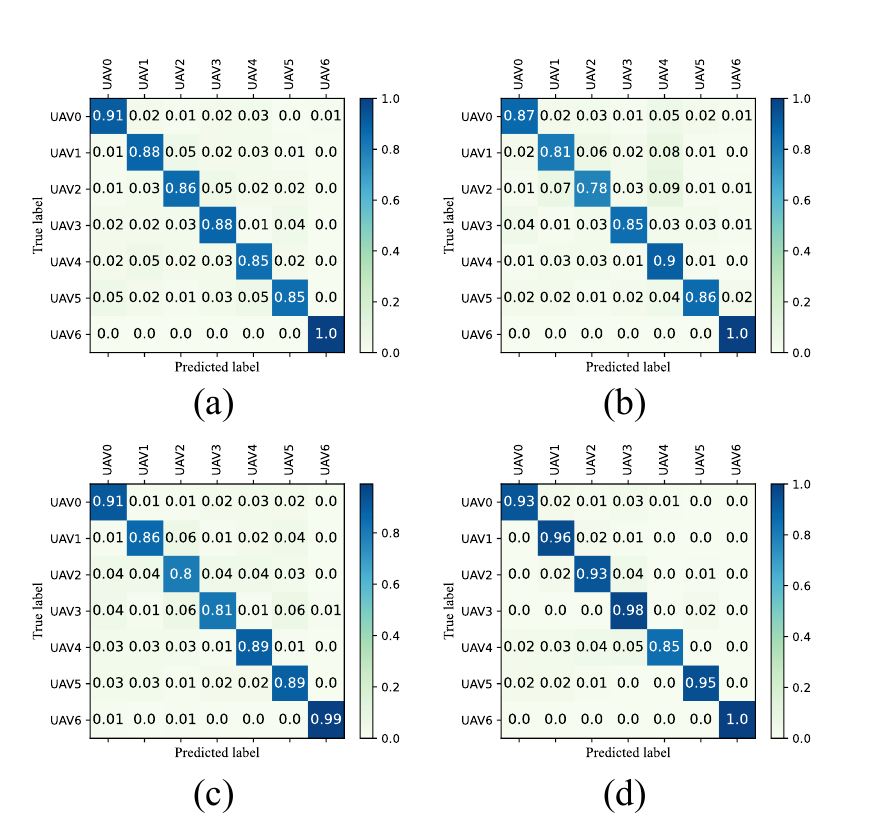} 
 \caption{Confusion matrices of the classification results evaluated at SNR = 0 dB. The subplots correspond to (a) HRank, (b) Sr-init, (c) PSR, and (d) Ours.}
 \label{fig:confusion_matrix}
\end{figure}

\subsection{Additional Experiments and Analysis}
\label{sec:Additional Experiments and Analyses}

As shown in \cref{fig:all_acc}, our method outperforms others more significantly in low-SNR environments. Therefore, we selected the specific case of SNR = $0$dB to conduct a detailed visualization analysis. By examining the feature representations and classification behaviors under this condition, we verify the robustness and effectiveness of HSCP.

\textbf{Feature Space Visualization.} \cref{fig:tsne} visualizes the t-SNE embeddings of feature representations extracted by different pruning methods at SNR = $0$dB. The comparison highlights the superior capability of HSCP in preserving the semantic manifold structure. As shown in \cref{fig:tsne} (b) and (c), the feature distributions from Sr-init and PSR exhibit severe class entanglement, where samples from different UAV categories are heavily mixed and lack clear decision boundaries. Although the Hrank method in \cref{fig:tsne}(a) improves local compactness, it still suffers from significant class overlap in the central region of the plot. In contrast, our method visualized in \cref{fig:tsne}(d) generates highly compact and well-separated clusters. The emergence of distinct inter-class edge suggests that HSCP effectively filters out redundant layers and channels, remaining competitive even under intense noise.

\textbf{Classification Performance Analysis.} We further certify these findings using the confusion matrices in \cref{fig:confusion_matrix} which quantify the class prediction precision at SNR=$0$dB. Our method demonstrates exceptional diagonal dominance compared to state-of-the-art baselines. An obvious example is the UAV3 category where HSCP achieves an accuracy of $0.98$. This significantly outperforms the HRank which attains $0.88$ and notably surpasses the PSR which drops to $0.81$. Similarly, for the UAV1 category, HSCP obtain an accuracy of $0.96$, marking a substantial improvement over the $0.86$ accuracy observed in the PSR. Furthermore, our approach effectively suppresses off-diagonal confusion. For instance, in the PSR, there is noticeable confusion between UAV1 and UAV2 where $0.06$ of UAV1 samples are misclassified as UAV2. In contrast, our method reduces this specific error to $0.02$. This demonstrates the superiority of HSCP in establishing a robust decision boundary, ensuring reliable identification in low-SNR environments.

\begin{table}[htbp]
  \centering
  \caption{Comparison of single-dimension pruning strategies versus our proposed joint framework on ResNet18.}
    \begin{tabular}{cccc}
    \toprule
    Pruning Strategy & Acc (\%) & Params (M) & FLOPs (M) \\
    \midrule
        Only Layer & 86.50  & 1.58  & 1038.83 \\
    Only Channel & 90.68 & 1.63  & 226.54 \\

    Layer + Channel & 91.62 & 1.52  & 241.58 \\
        Layer + Channel + Mixup & 94.11 & 1.52  & 241.58 \\
    \bottomrule
    \end{tabular}%
  \label{tab:pruning_strategy}%
\end{table}%

\begin{table}[htbp]
  \centering
  \caption{Impact of the $\alpha$ on the top-1 Accuracy of the Pruned ResNet18.}
    \begin{tabular}{cccccc}
    \toprule
    Alpha & 0     & 0.3   & 0.5   & 0.7   & 0.9 \\
    \midrule
    Acc(\%) & 92.62 & 95.71 & 95.62 & 96.14 & 96.02 \\
    \bottomrule
    \end{tabular}%
  \label{tab:alpha}%
\end{table}%

\begin{table}[t]
  \centering
  \caption{Comparison of Computational Efficiency and Accuracy between SOTA Hand-crafted Architectures and HSCP.}
    \begin{tabular}{c|cccc}
    \toprule
    \multicolumn{1}{c}{Methodology} & Approach & Params & Flops & Acc(\%) \\
    \midrule
    \multirow{2}[2]{*}{Custom} & Zhou et al.~\cite{zhou2025lightweight} & $4.00\times10^4$ & $1.80\times10^8$ & 95.79 \\
          & Cai et al.~\cite{cai2024toward} & $1.27\times10^6$ & -     & 93.25 \\
    \midrule
    Compressed & Ours & $5.02\times10^5$ & $6.28\times10^6$ & 95.64 \\
    \bottomrule
    \end{tabular}%
  \label{tab:compare}%
\end{table}%

\subsection{Ablation Study}
\label{sec: Ablation Study}
\textbf{Impact of Different Pruning Strategies.} To verify the necessity of hierarchical pruning, we conduct an ablation study on ResNet18 to isolate the contributions of layer and channel pruning. As shown in \cref{tab:pruning_strategy}, applying either strategy in isolation results in a sharp performance drop. Specifically, the \textit{Only Layer} strategy retains a high computational burden and results in a severe accuracy drop to $86.50\%$. This indicates that removing layers alone cannot effectively eliminate redundancy in the width dimension. Similarly, \textit{Only Channel} leads to an accuracy reduction to $90.68\%$. In contrast, our \textit{Layer + Channel} achieves the minimum parameters while maintaining a higher accuracy of $91.62\%$. Moreover, Mixup training further improves the accuracy to $94.11\%$, which indicates that our method effectively balances efficiency and performance.

\textbf{Impact of Mixup Strength.} We further investigate the sensitivity of the pruned ResNet18 model to the hyperparameter $\alpha$ as summarized in \cref{tab:alpha}. When $\alpha$ is fixed at 0, the training process follows the standard way without sample blending and produces a baseline accuracy of $92.62\%$. Introducing the mixing mechanism by increasing $\alpha$ to $0.3$ immediately yields a substantial performance gain where the accuracy rises to $95.71\%$. Crucially, this performance improvement remains robust across a wide range of parameter values. As $\alpha$ is adjusted to $0.5$ and $0.7$, the accuracy stabilizes at a high level of $95.62\%$ and $96.14\%$ respectively, significantly outperforming the baseline. Meanwhile $\alpha=0.9$ also maintains a competitive accuracy of $96.02\%$. The results collectively indicate that HSCP is not overly sensitive to specific $\alpha$ tuning within this effective range. Given the consistent gains and the balance between signal preservation and augmentation intensity, we adopted the moderate value of $0.5$ as the default configuration for all experiments to ensure stable and generalizable training.

\textbf{Comparison with Hand-crafted Architectures.} To further validate the superiority of HSCP, we compare it with the SOTA manually designed networks including Zhou et al.~\cite{zhou2025lightweight} and Cai et al.~\cite{cai2024toward} as shown in Table \ref{tab:compare}. Specifically, although Zhou et al. achieves a comparable accuracy of $95.79\%$ with fewer parameters, it suffers from a high computational burden of $1.80 \times 10^8$ FLOPs. In contrast, our method reduces the computational cost to $6.28 \times 10^6$ FLOPs which is approximately $28$ times more efficient than Zhou et al. while maintaining a high accuracy of $95.64\%$. Furthermore, compared to Cai et al., our method achieves a higher accuracy with an improvement of $2.39\%$ while utilizing fewer parameters, specifically reducing the count from $1.27 \times 10^6$ to $5.02$. This demonstrates that HSCP achieves a better trade-off between computational complexity and performance than manual designs.

\section{Conclusion}
\label{sec:Conclusion}
In this paper, we propose HSCP, a hierarchical spectral clustering pruning framework that combines layer pruning with channel pruning to achieve extreme compression, high performance, and efficient inference. Specifically, HSCP consists of two stages. In the first stage, spectral clustering based on CKA similarity is applied to layer representations to achieve layer pruning. Then, the unified spectral analysis strategy utilizes channel similarity to diagnose redundancy and removes these channels. Experiments on the UAV-M100 dataset demonstrate that HSCP outperforms existing channel and layer pruning methods while achieving superior robustness in low SNR ratio environments.

\bibliographystyle{IEEEtran}
\bibliography{reference}

\begin{IEEEbiography}[{\includegraphics[width=1in,height=1.25in,clip,keepaspectratio]{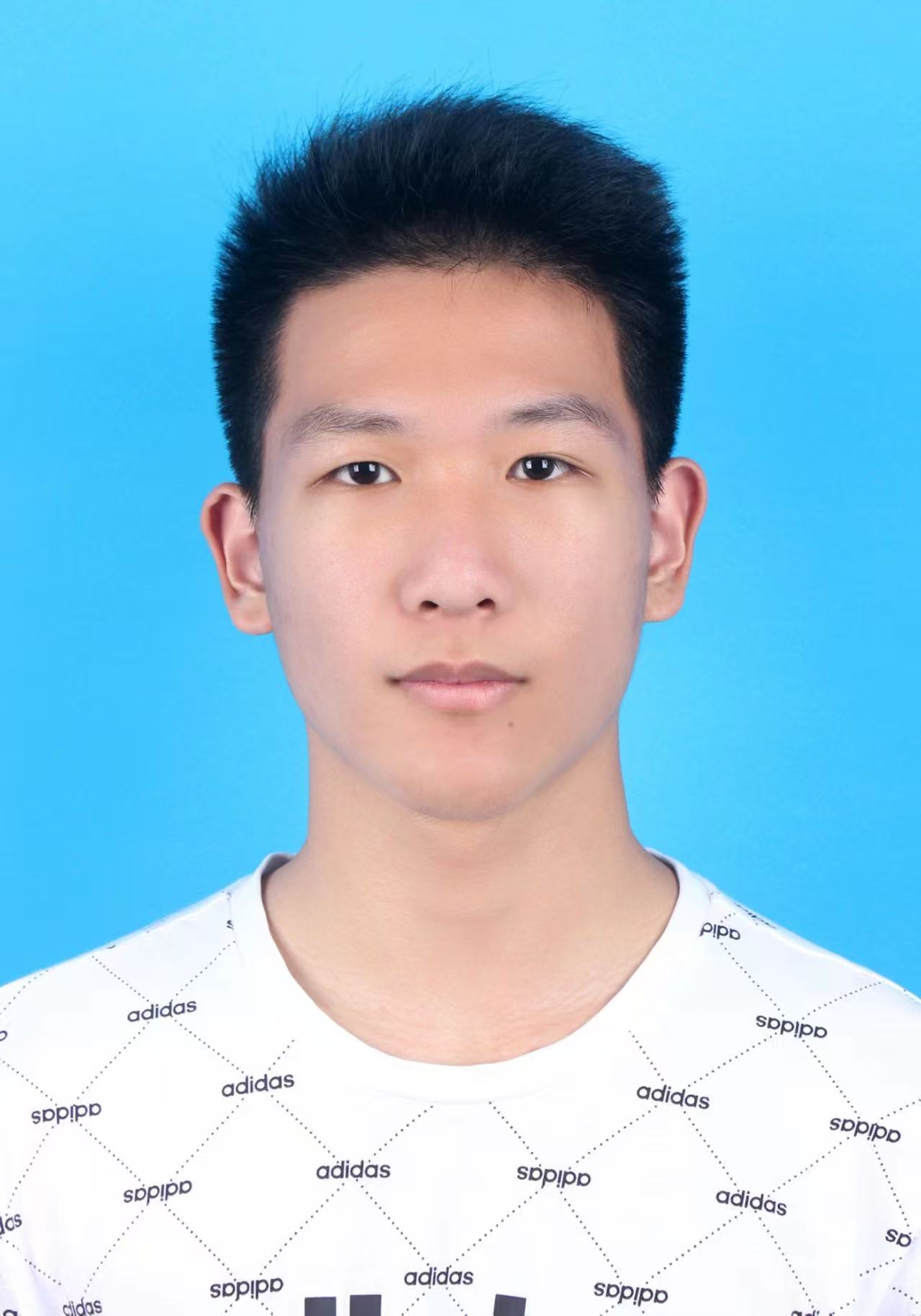}}]{Maoyu Wang}
is currently pursuing the B.S. degree at the College of Computer Science and Technology, Zhejiang University of Technology, Hangzhou, China. His research interests include deep learning and computer vision.
\end{IEEEbiography}

\begin{IEEEbiography}[{\includegraphics[width=1in,height=1.25in,clip,keepaspectratio]{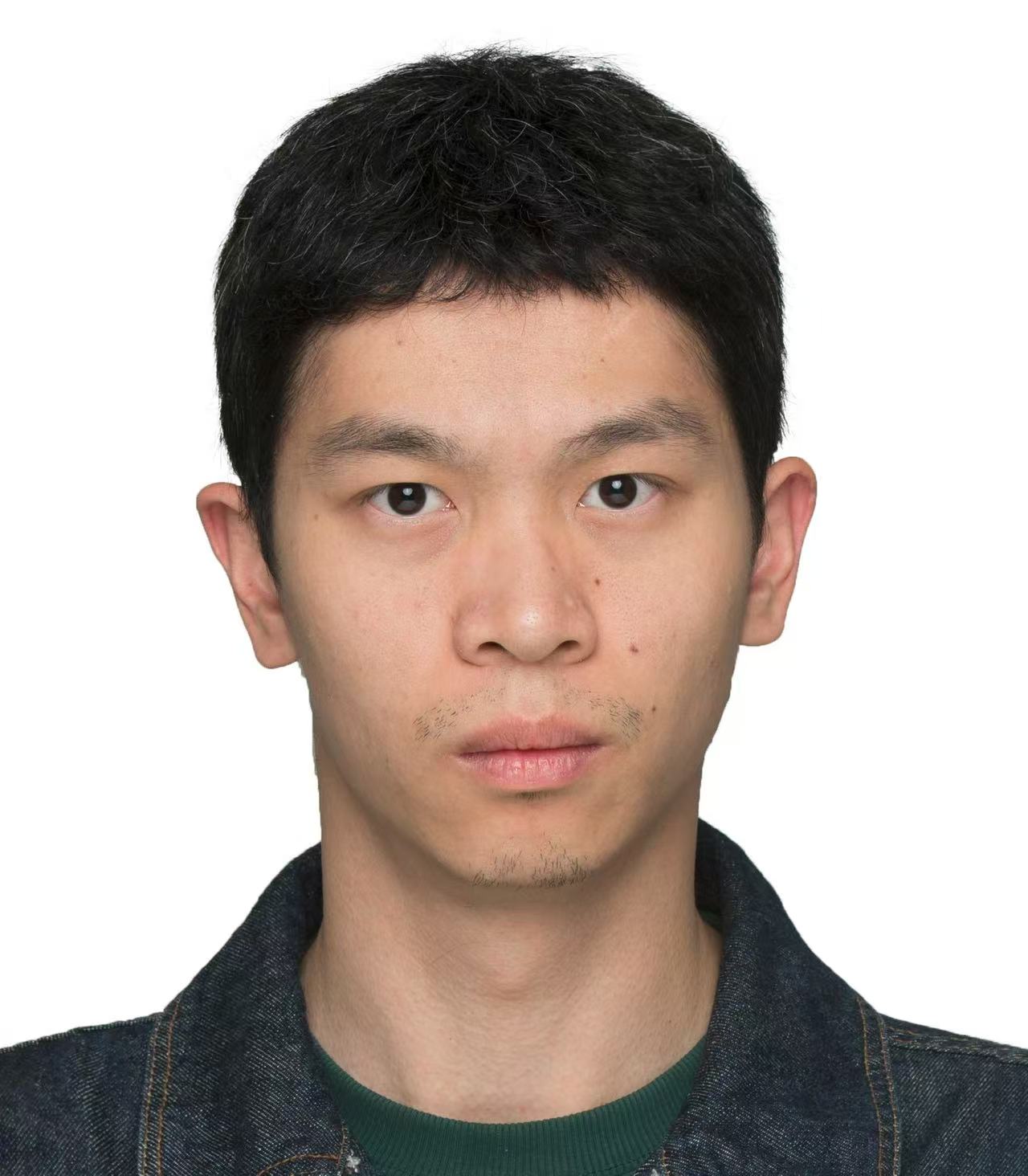}}]{Yao Lu}
received his B.S. degree from Zhejiang University of Technology and is currently pursuing a Ph.D. in control science and engineering at Zhejiang University of Technology. He is a visiting scholar with the Centre for Frontier AI Research, Agency for Science, Technology and Research, Singapore. He has published several academic papers in international conferences and journals, including ECCV, AAAI, TNNLS, Neurocomputing and TCCN. He serves as a reviewer of ICLR, CVPR, ICCV, NeurIPS and TNNLS. His research interests include deep learning and computer vision, with a focus on artificial intelligence and model compression.
\end{IEEEbiography} 

\begin{IEEEbiography}[{\includegraphics[width=1in,height=1.25in,clip,keepaspectratio]{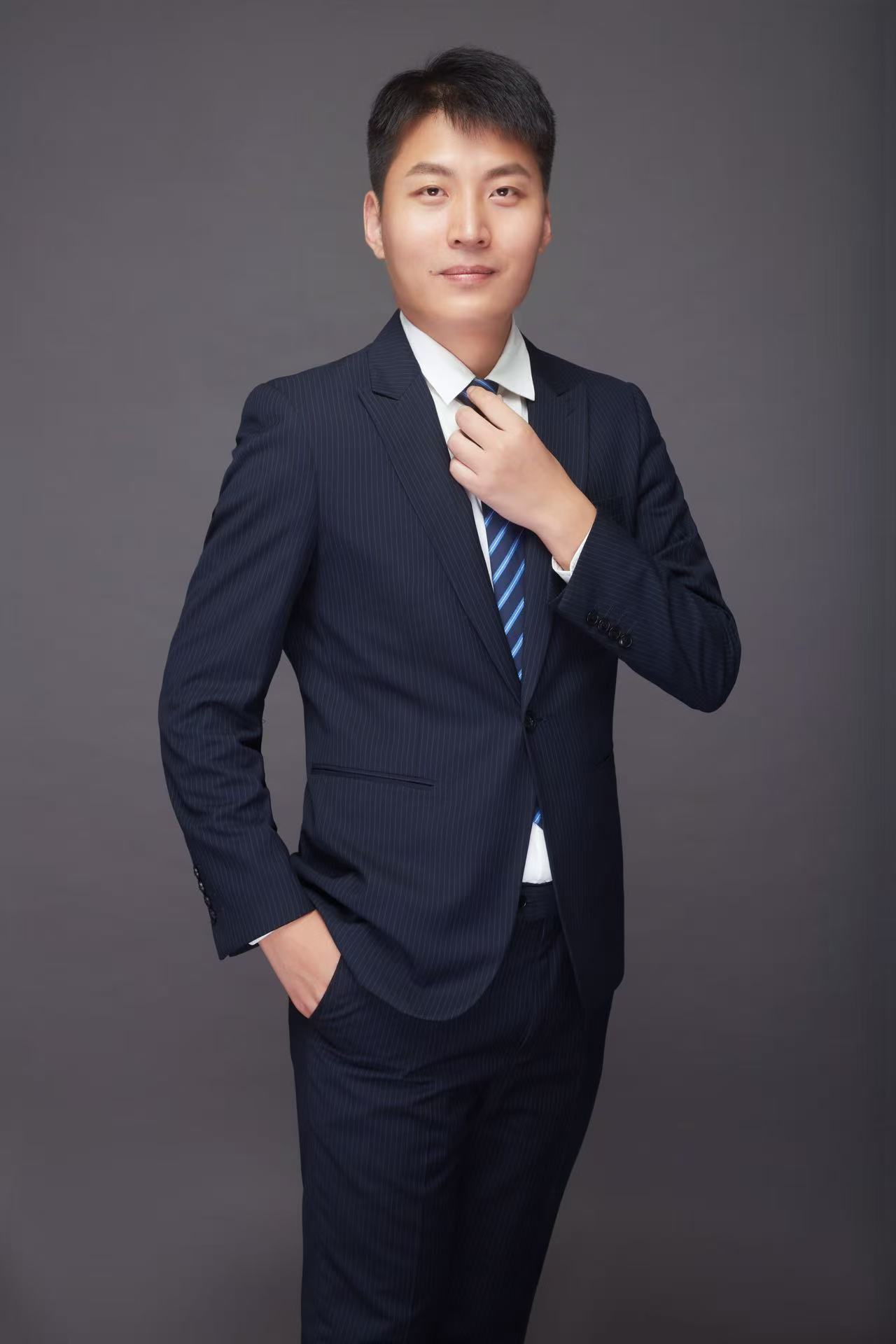}}]{Bo Zhou}
received the B.S. degree from the Ocean College, Zhejiang Ocean University, Zhoushan, China, in 2011, the M.S. degree from the Ocean College, Zhejiang University, Hangzhou, China, in 2014 and the Ph.D. degree with the Institute of Cyberspace Security, College of Information Engineering, Zhejiang University of Technology, Hangzhou, China, in 2024.

He is currently a Lecturer with the Department of Intelligent Control, Zhejiang Institute of Communications, Hangzhou. His research is about the node
importance of social networks, machine learning, and complex network.
\end{IEEEbiography}

\begin{IEEEbiography}[{\includegraphics[width=1in,height=1.25in, clip,keepaspectratio]{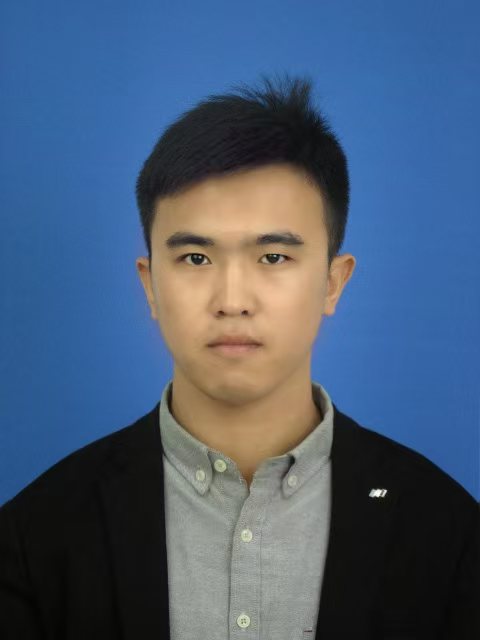}}]{Zhuangzhi Chen}
received the B.S. and Ph.D. degrees in Control Theory and Engineering from Zhejiang University of Technology, Hangzhou, China, in 2017 and 2022, respectively. He was a visiting scholar with the Department of Computer Science, University of California at Davis, CA, in 2019. He completed his postdoctoral research in Computer Science and Technology at Zhejiang University of Technology, in 2024. He is currently working as an associate researcher at the Cyberspace Security Research Institute of Zhejiang University of Technology, and also serving as the deputy director and associate researcher at the Research Center of Electromagnetic Space Security, Binjiang Institute of Artificial Intelligence, ZJUT, Hangzhou 310056, China.

His current research interests include deep learning algorithm, deep learning-based radio signal recognition, and security AI of wireless communication.
\end{IEEEbiography} 


\begin{IEEEbiography}[{\includegraphics[width=1in,height=1.25in,clip,keepaspectratio]{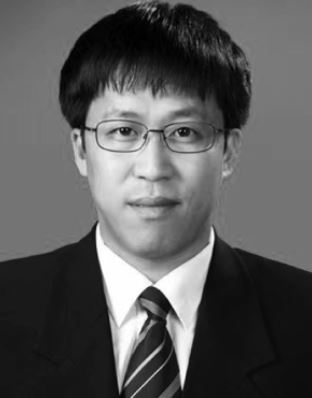}}]{Yun Lin}
(Member, IEEE) received the B.S. degree in electrical engineering from Dalian Maritime University, Dalian, China, in 2003, the M.S. degree in communication and information system from the Harbin Institute of Technology, Harbin, China, in 2005, and the Ph.D. degree in communication and information system from Harbin Engineering University, Harbin, in 2010. From 2014 to 2015, he was a Research Scholar with Wright State University, Dayton, OH, USA. He is currently a Full Professor with the College of Information and Communication Engineering, Harbin Engineering University. He has authored or coauthored more than 200 international peer-reviewed journal/conference papers, such as IEEE Transactions on Industrial Informatics, IEEE Transactions on Communications, IEEE Internet of Things Journal, IEEE Transactions on Vehicular Technology, IEEE Transactions on Cognitive Communications and Networking, TR, INFOCOM, GLOBECOM, ICC, VTC, and ICNC. His current research interests include machine learning and data analytics over wireless networks, signal processing and analysis, cognitive radio and software-defined radio, artificial intelligence, and pattern recognition.
\end{IEEEbiography}

\begin{IEEEbiography}[{\includegraphics[width=1in,height=1.25in,clip,keepaspectratio]{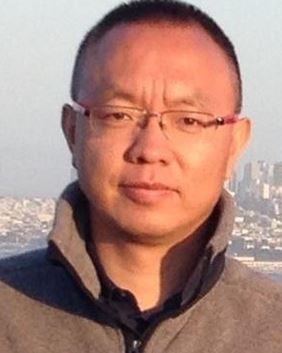}}]{Qi Xuan}
(Senior Member, IEEE) received the B.S. and Ph.D. degrees in control theory and engineering from Zhejiang University, Hangzhou, China, in 2003 and 2008, respectively. He was a Postdoctoral Researcher with the Department of Information Science and Electronic Engineering, Zhejiang University from 2008 to 2010, and a Research Assistant with the Department of Electronic Engineering, City University of Hong Kong, Hong Kong, in 2010 and 2017, respectively. From 2012 to 2014, he was a Postdoctoral Fellow with the Department of Computer Science, University of California at Davis, Davis, CA, USA. He is currently a Professor with the Institute of Cyberspace Security, College of Information Engineering, Zhejiang University of Technology, Hangzhou, and also with the PCL Research Center of Networks and Communications, Peng Cheng Laboratory, Shenzhen, China. He is also with Utron Technology Company Ltd., Xi’an, China, as a Hangzhou Qianjiang Distinguished Expert. His current research interests include network science, graph data mining, cyberspace security, machine learning, and computer vision.
\end{IEEEbiography}

\begin{IEEEbiography}[{\includegraphics[width=1in,height=1.25in,clip,keepaspectratio]{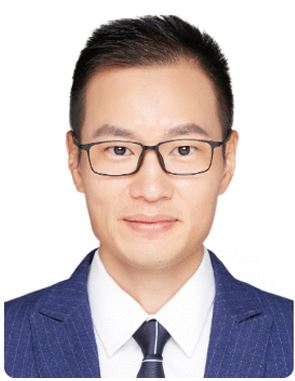}}]{Guan Gui}
    received the Dr. Eng. degree in information and communication engineering from the University of Electronic Science and Technology of China, Chengdu, China, in 2012. From 2009 to 2012, with financial support from the China Scholarship Council and the Global Center of Education, Tohoku University, he joined the Wireless Signal Processing and Network Laboratory (Prof. Adachis laboratory), Department of Communications Engineering, Graduate School of Engineering, Tohoku University, as a Research Assistant and a Post Doctoral Research Fellow, respectively. From 2012 to 2014, he was supported by the Japan Society for the Promotion of Science Fellowship as a Post Doctoral Research Fellow with the Wireless Signal Processing and Network Laboratory. From 2014 to 2015, he was an Assistant Professor with the Department of Electronics and Information System, Akita Prefectural University. Since 2015, he has been a Professor with the Nanjing University of Posts and Telecommunications, Nanjing, China. He is currently involved in the research of big data analysis, multidimensional system control, super-resolution radar imaging, adaptive filter, compressive sensing, sparse dictionary designing, channel estimation, and advanced wireless techniques. He received the IEEE International Conference on Communications Best Paper Award in 2014 and 2017 and the IEEE Vehicular Technology Conference (VTC-spring) Best Student Paper Award in 2014. He was also selected as a Jiangsu Special Appointed Professor, as a Jiangsu High-Level Innovation and Entrepreneurial Talent, and for 1311 Talent Plan in 2016. He has been an Associate Editor of the Wiley Journal Security and Communication Networks since 2012.
\end{IEEEbiography}

\clearpage

\vfill

\end{document}